# AI Stories: An Interactive Narrative System for Children

**Ben Burtenshaw**
Computational Linguistics & Psycholingustics
The University of Antwerp
Antwerp, 2000, Belgium
benjamin.burtenshaw@uantwerpen.be

### Abstract

AI Stories is a proposed interactive dialogue system, that lets children co-create narrative worlds through conversation. Over the next three years this system will be developed and tested within pediatric wards, where it offers a useful resource between the gap of education and play. Telling and making stories is a fundamental part of language play, and its chatty and nonsensical qualities are important; therefore, the prologued usage an automated system offers is a benefit to children. In this paper I will present the current state of this project, in its more experimental and general guise. Conceptually story-telling through dialogue relates to the pre-print interpretation of story, beyond the static and linear medium, where stories were performative, temporal, and social.

## Introduction

AI Stories is an applied Computational Creativity and Natural Language Processing project to develop an interactive dialogue system for children. The eventual system will be applied in the health and education sectors, autonomously generating stories through dialogue. Here I will present the current state of that four year project.

Language play through dialogue is a fundamental process in the development of a child's language skills, through the creation of fictional worlds children learn how to use language. A key component to language play is interactive narrative; the process of making a story through dialogue and questioning. Interactive narrative places stories into the temporal and social space of dialogue, where children act as meaning makers, intuitively learning the building blocks of language (Wells 1986). Children experience a rich and changing space of narrative, where they can actively engage in their own interests. AI Stories would encase this established process within a usable and friendly software.

Anyone that's witnessed children playing with interfaces like Siri probably knows that they are an exciting and appealing innovation. Children will often ask nonsensical questions which reflect a creative interest in broader topics; for Example, "How many people in the Earth have mommy's?". Though effort has been made to give Siri an air of awareness through one-off comebacks to queries like "Can I have a bedtime story?", with "Next you'll be asking me for a glass of milk. And a dark matter cookie." There is a clear lack of expanded interaction in a way that encourages the child's enthusiasm. The dialogue could build upon the child's nonsensical queries in a way that deepens their exploration. Furthermore, a diachronic approach allows for online information to enrich the story rather than end it. For example, by retrieving information from wikipedia, the dinosaur in our a story could be "a herbivore named Massy, short for Massospondylidae". This applies the very natural process of articulating information through narrative, which is commonplace in children's education.

## Related Work

Narrative generation has been a central topic of computational creativity for decades. One of the first examples is Tale-Spin, a system that generates Aesop's Fables guided by the user's input (Meehan 1977). Universe by Michaeal Lebowitz draws on a database of character definitions, plot outlines, and dialogues, to weave together new stories; however, such systems are limited by their fixed content (Lebowitz 1983). Callaway and Lester produced Storybook, a narrative prose generator that identifies the narrative phases in a text, and uses them to generate a new narrative (Callaway and Lester 2002); however, systems of this kind tend to reproduce narrative tropes, and can become boring over prologued useage (Swanson and and Gordon 2008). McIntyre and Lapata developed one of the first search-based narrative systems that uses user-input to search a database for relating phrases (McIntyre and Lapata 2009). This architecture produces interesting relationships between phrases, but is to sporadic to create longer narratives. To counter these semantic fluctuations, Riedl and Bultiko use the dialogue history to guide the generated text (Riedl and Bulitko 2012); however, by their own admis-sion, this becomes monotonous to the user, with little room for surprise past their own input.

Interactive Narrative generation is also well established within computational creativity (Dehn 1981; Klein and Balsiger 1973; Rafael Pérez Ý Pérez and Sharples 2001). Researchers have used intermittent feedback from the user to inform a logical and natural narrative. Unlike a one-off narrative generation, where a system must string together a series of sentences by itself, interactive systems can adapt each sentence to the user's input. For a thorough review of early interactive narrative systems see Gervas 2009. User input comes with it's own challenges, as the possible responses for most dialogues are huge, and the system needs to incorporate these in a way that both appeases the user and maintains trajectory. Narrative intelligence is the ability to craft and communicate stories, recent research has improved the application of narrative intelligence (Riedl 2016); systems are able to respond to queries about the narrative of a text, and reorder sentences into a narrative order.

# Method

A completed AI Stories system will use dialogue to create a congruent and multi-tonal story, and draw on online information to enrich the narrative. Here I will propose the current state of that system.

Interactive narratives do not need to be straightforward nor logical, human authors often use narrative shifts to tell interesting stories. That said, certain elements need to remain constant, and a computational system should respect these; for example, the description of characters and locations. If these shift throughout the story a user can easily become confused. However, elements like the structure can be more fluidic and open to speculation.

Anyone that has told stories to children, knows how challenging their interjections can be. One has to respond in a way that maintains their interest, but must also maintain the narrative trajectory. In dialogue we weave together the threads of narrative in a way that both respects interjection and remains consistent. Either focus alone produces sporadic or boring conversation, yet combined they allow us to create complex stories.

## Subsystems

Emulating this multi-tonal approach to dialogue, the system proposed here uses subsystems to adversarially generate responses to the user's input. These proposed responses are then considered by a selector system that chooses the best response. The eventual system could use a range subsystems, but here I will discuss using three; a topic based question answering system, a context based sequence to sequence system, and a template based poetry system.

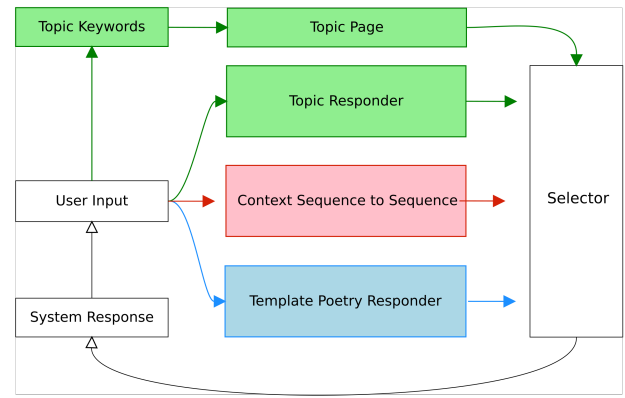

## Topic Based System

The topic based system treats the user's input as queries relating to a predefined topic text; for example, a wikipedia page about dinosaurs. The system is given the topic as a search term, which it uses to collect the page. It then searches the page for answers relating to the user input. The system returns its proposed response and a certainty score to the selector.

## Context Based System

The context based system uses a sequence to sequence neural network to respond to the input of the user. Each user input and actual output build a context memory which it uses to form a response. This approach uses techniques from neural machine translation, and have had proven success responding on single turn dialogue (Wu et al. 2016; C. Li et al. 2016; J. Li et al. 2016; Xing et al. 2016).

## Poetry and Humor System

The poetry based system uses template sentences, rhyme, and word definitions to generate humorous and entertaining responses. This relates to the established work of researchers like Pablo Gervas, Tony Veale, and Simon Colton.

The poetry and humor system is the default choice for the selector. An applied system would rely on poetry and humor to maintain the user's interest; more homogeneous systems can fail at the sporadic musings of a child, revealing a lack in the system's comprehension. However, we can learn here from Woody Allen's character in the 1977 film Manhattan, who in reference to his own TV sitcom says "It's worse than not insightful: it's not funny", implying that if one can't be insightful they should at least be funny.

## The Selector

Alone the sub-systems would lead to repetitive or monotonous conversation. However, brought together they are able to individually contribute some of the key building blocks of dialogue. The selector's task is to decide which of the proposed sentences would maintain the conversation, it does this using a combination of reinforcement learning and lexical analysis. The proposals from the subsystems are encoded as vector representations, and reinforcement learning is used to train a policy that optimizes toward conversation with longer consistency. I have used Q-learning because it builds a Q table that works well alongside hand coded lexical rules. The reinforcement learning task is framed as a simplified Partially Observable Markov Decision Process (S,A,R); where S is a set of previous dialogue turns, A are the proposed sentences from the subsystems, and R is a reward score. Choosing a sentence will provide a reward R(s,a) based on the similarity of that sentence to the following one. The goal of the selector is to choose the sentence which will maximize the reward function by being closest to the next sentence in training. The selector is trained on dialogue and TV subtitle corpora, which allows the selector to draw on offline training and choose responses that emulate human conversation patterns. The hard coded lexical analysis can then block or privilege proposals based on the lexical quality of the input sentence. For example, if a proposed response is offensive the selector can reject it. Likewise, if an input statement relates to a specific system's purpose; like "Tell me a joke about Dinosaurs", or "Do all Dinosaurs have legs?", the selector can weight the appropriate system so that it is more likely to be used.

## Results

The system and implementation discussed here are an early proposal. That said, the use of subsystems produces interesting forms of dialogue that reflect the multi-tonal tendencies of human conversation. The results at this stage are not enough to be called dialogue nor narrative, but their quality reflects the dialogue that will be fundamental to autonomous conversations with children. Furthermore, the combination of hand coded lexical treatment of turns with neural networks, presents a novel and powerful application of the two within a children's dialogue context.

## Future Work

AI Stories will be expanded over the next 3 years to include sub-systems that reflect more diverse tones in dialogue, and the selector will be complexified to discriminate more subtle distinctions in language. For example, by incorporating a noisy interpretation of the current state the selector would be able to better reflect the dialogue history, and a factor for the likelihood of each sentence to produce longer dialogues would allow the selector to better mimic human conversation patterns.

## Conclusion

In this paper I have presented the current state of my PhD project, which is based on the system AI Stories. The intended system would use dialogue to produce narratives for children, and could be left to autonomously guide children through educational content.